\documentclass[10pt, a4paper]{article}

\usepackage{lrec-coling2024} % this is the new style
\usepackage{amsmath,amsthm,amsfonts,amssymb,amscd}
\usepackage{mathrsfs}
\usepackage{times}
\usepackage{graphicx} 
\usepackage{wrapfig}
\usepackage{subcaption}
\usepackage{tabularx} % 引入tabularx包
\usepackage{booktabs} % 用于更优雅的表格线条
\usepackage{multirow} % 用于跨行
\usepackage{arydshln}
\usepackage{array}
\usepackage{rotating}
\usepackage{pgfplots} 
\usepackage{tikz}
\usetikzlibrary{shapes.geometric, arrows, positioning}
\usetikzlibrary{positioning, fit, calc}

\title{Llama-VITS: Enhancing TTS Synthesis with Semantic Awareness}
\name{Xincan Feng\textsuperscript{\dag \ddag}, Akifumi Yoshimoto\textsuperscript{\ddag}} 

\address{\textsuperscript{\dag}NARA Institute of Science and Technology, \textsuperscript{\ddag}CyberAgent Inc\\
         feng.xincan.fy2@is.naist.jp, yoshimoto\_akifumi\_xa@cyberagent.co.jp\\}

\abstract{
Recent advancements in Natural Language Processing (NLP) have seen Large-scale Language Models (LLMs) excel at producing high-quality text for various purposes. Notably, in Text-To-Speech (TTS) systems, the integration of BERT for semantic token generation has underscored the importance of semantic content in producing coherent speech outputs. Despite this, the specific utility of LLMs in enhancing TTS synthesis remains considerably limited. This research introduces an innovative approach, Llama-VITS, which enhances TTS synthesis by enriching the semantic content of text using LLM. Llama-VITS integrates semantic embeddings from Llama2 with the VITS model, a leading end-to-end TTS framework. By leveraging Llama2 for the primary speech synthesis process, our experiments demonstrate that Llama-VITS matches the naturalness of the original VITS (ORI-VITS) and those incorporate BERT (BERT-VITS), on the LJSpeech dataset, a substantial collection of neutral, clear speech. Moreover, our method significantly enhances emotive expressiveness on the EmoV\_DB\_bea\_sem dataset, a curated selection of emotionally consistent speech from the EmoV\_DB dataset, highlighting its potential to generate emotive speech.
 \\ \newline  }

\begin{document}

\maketitleabstract

\section{Introduction}
% 文本到语音（TTS）合成是将书面文本转换为相应的口语形式，以增加内容的可访问性的技术。此项技术可以应用于有声书、虚拟助手的制作。然而，由于传统的TTS模型在处理输入的文本时仅考虑了文本的声学特征，因此虽然可以生成自然流畅的语音，但在理解文本中的语义和情感信息时有明显不足。
% 随着自然语言处理（NLP）的显著进步，尤其是语言模型（LM），比如双向编码器表示从变压器（BERT）和生成式预训练（GPT），在各种自然语言理解和表达的任务中展现的强大能力，研究者们提出了各种基于BERT的TTS模型，以提高生成的语音的语义表达能力。但是，由于BERT 模型的参数规模较小，而且需要设计适当的微调任务才可提升能力，因此这类模型在不同应用中的有效性和灵活性受到限制。
% 另一方面，大规模语言模型（LLMs），如 Llama2，不仅计算资源需求越来越小、文本生成水平越来越高，还天然具有出色的零样本学习能力，此外，只需要对极少量的参数进行提示调整也可以达到和微调一样的改善性能的效果。然而，这类大语言模型对TTS任务的潜力还未得到研究。
% 鉴于上述背景，我们提出了Llama-VITS，该模型在传统模型的基础上利用了由Llama2模型提取的输入文本的语义表示，生成的语音不仅保留了文本的声学信息，而且可以理解和表达语义和情感。通过充分的客观和主观评估，证实了我们的模型优于没有语义输入或基于BERT的TTS基线模型。
% 本工作的主要贡献总结如下：
% - 我们提出了一个可以利用 Llama2 的语义理解和表达能力的 TTS 模型，与基线模型相比，我们的模型不仅具有同等或更好的声学表现，还具有显著更好的理解和表达语义和情感的能力。据我们所知，之前没有研究过GPT-like的模型在 TTS 模型中的应用。
% - 通过实证分析，我们展示了Llama-VITS中全局话语级令牌比序列令牌提供了更多的改进，这与基于BERT的 TTS模型中所观察的现象恰好相反。
% - 我们提供了一个新的评估框架，不仅包含了传统的合成语音的声学表现的客观指标，还增加了 ESMOS 这个主观指标以评价情感表现。
% 我们的代码、预训练模型、音频样本和过滤后的情感数据集EmoV\_DB\_bea\_sem在xxxx开放。

Text-to-Speech (TTS) synthesis is a technology that transforms written text into its spoken equivalent, thereby enhancing content accessibility. This technology finds application in the production of audiobooks~\citep{chen2022unsupervised} and virtual assistants~\citep{wu2023simmc}. However, traditional TTS models, which primarily focus on the acoustic features, often fall short in comprehending the semantic and emotional information embedded within the text.

With the significant advancements in Natural Language Processing (NLP) technologies, particularly through Language Models (LMs) such as BERT~\citep{devlin2019bert} and GPT~\citep{radford2018improving, brown2020language}, which have demonstrated formidable capabilities in understanding and generating natural language, researchers have proposed various BERT-based TTS models~\citep{mukherjee2022Text, abbas2022expressive, li2023phonemelevel, guo2022PromptTTS} to improve the expressiveness of synthesized speech. Nonetheless, the effectiveness and flexibility of BERT-based TTS models in diverse applications are limited due to the smaller parameter size of BERT models and the necessity for designing specific fine-tuning tasks to enhance their capabilities.

On the other hand, Large-scale Language Models (LLMs), such as Llama2~\citep{touvron2023llama}, not only require decreasing computational resources and achieve higher levels of text generation but also possess excellent zero-shot learning capabilities. Moreover, they can achieve improvements comparable to fine-tuning by adjusting only a minimal number of parameters through prompt tuning~\citep{liu2022ptuning, tu2022prompt}. However, the potential of these LLMs for TTS tasks has not been fully explored.

In light of this context, we introduce Llama-VITS, a model that leverages semantic representations extracted from Llama2 on top of a state-of-the-art TTS model, VITS \citep{kim2021conditional}, enabling the generated speech to retain acoustic information while understanding and expressing semantics and emotions. Through comprehensive objective and subjective evaluations, Llama-VITS has been verified to surpass TTS baselines without semantic input or those integrated with BERT.

\begin{figure*}[t]
    \centering
    \includegraphics[width=\linewidth]{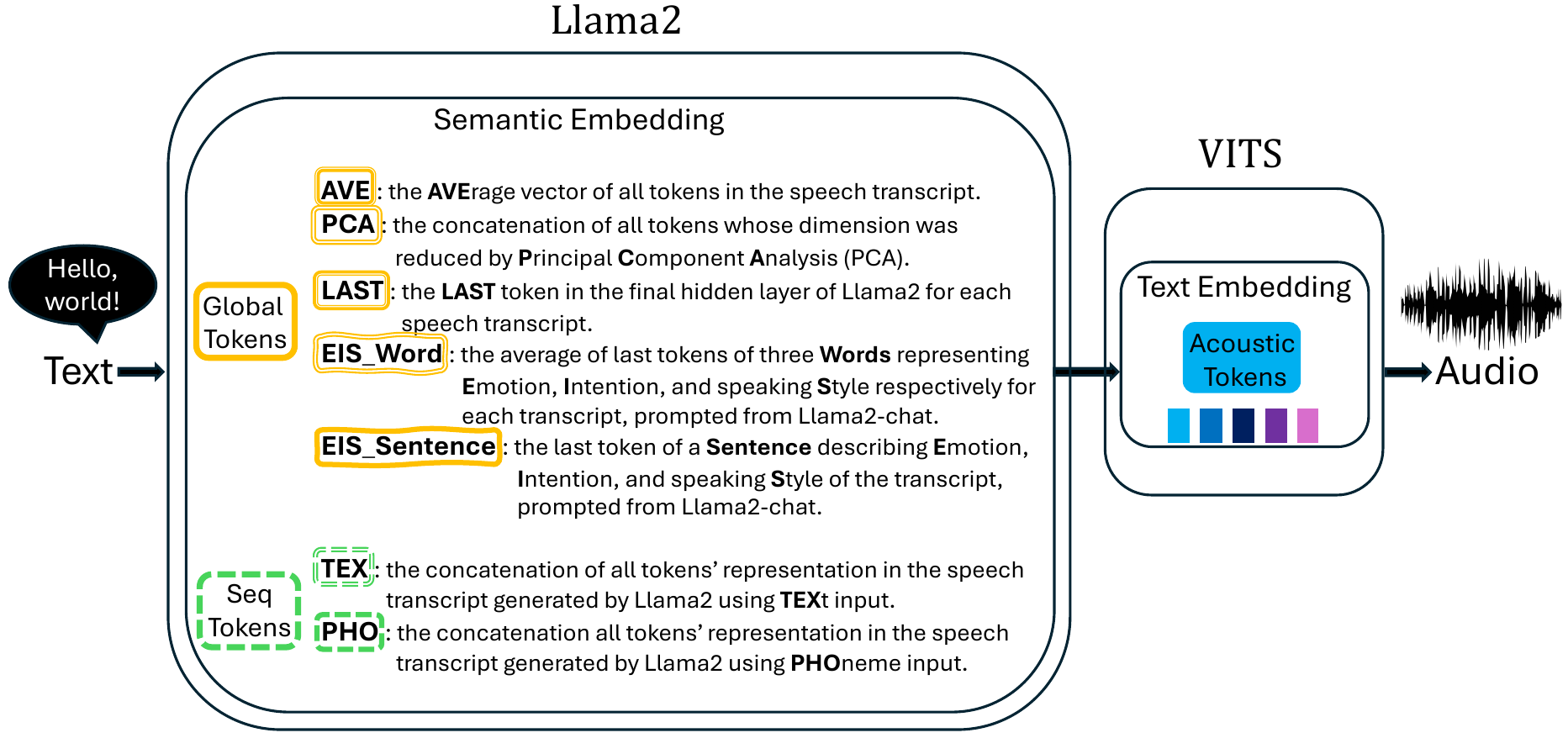}
    \caption{Pipeline of Llama-VITS}
    \label{fig:pipeline}
\end{figure*}
\begin{figure*}[t]
    \centering
    \begin{subfigure}[b]{0.49\linewidth}
        \includegraphics[width=\linewidth]{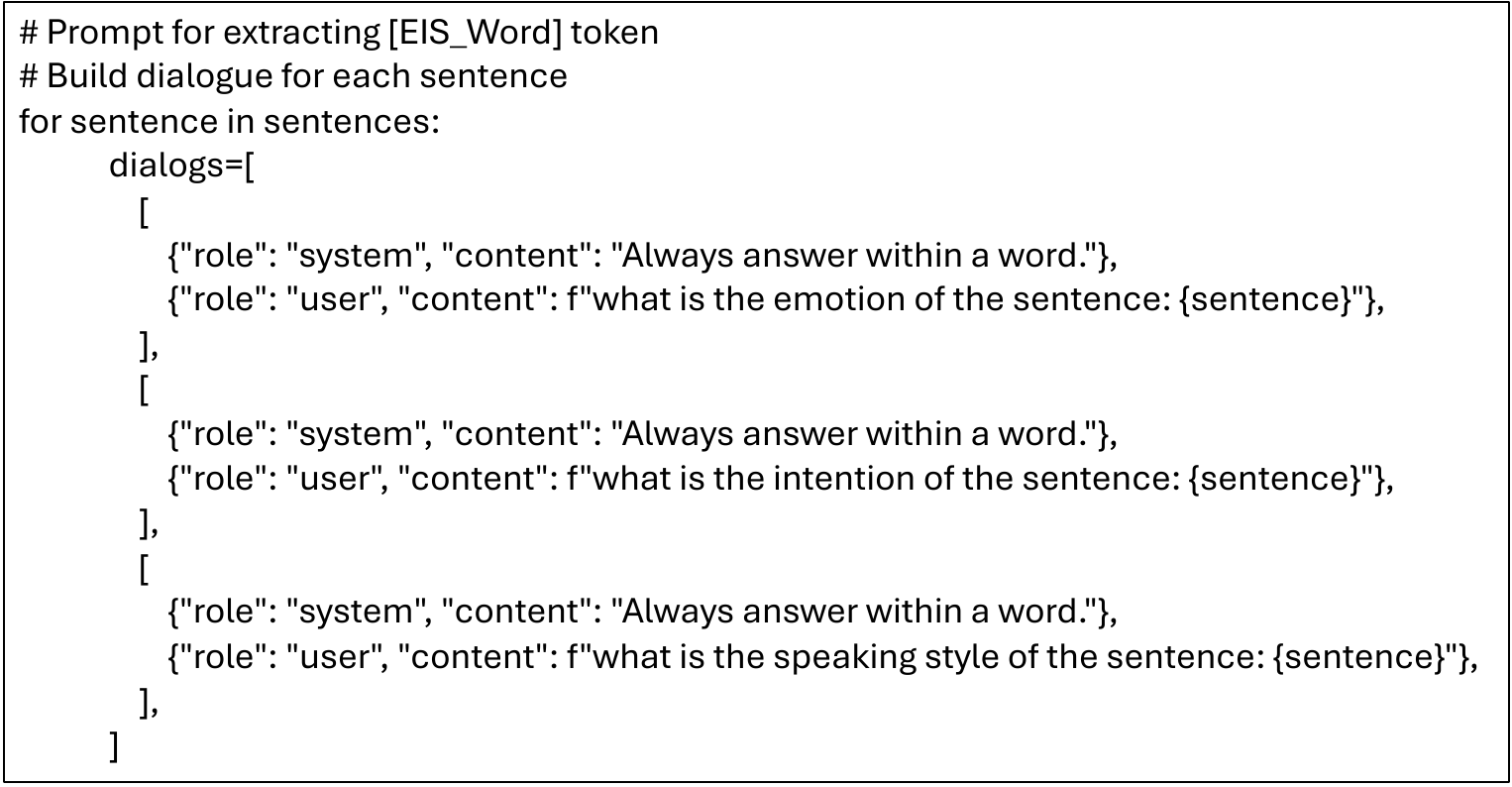}
        \caption{Prompts used for extracting [EIS\_Word] tokens}
        \label{fig:prompt_word}
    \end{subfigure}
    \hfill % This adds a little space between the subfigures
    \begin{subfigure}[b]{0.49\linewidth}
        \includegraphics[width=\linewidth]{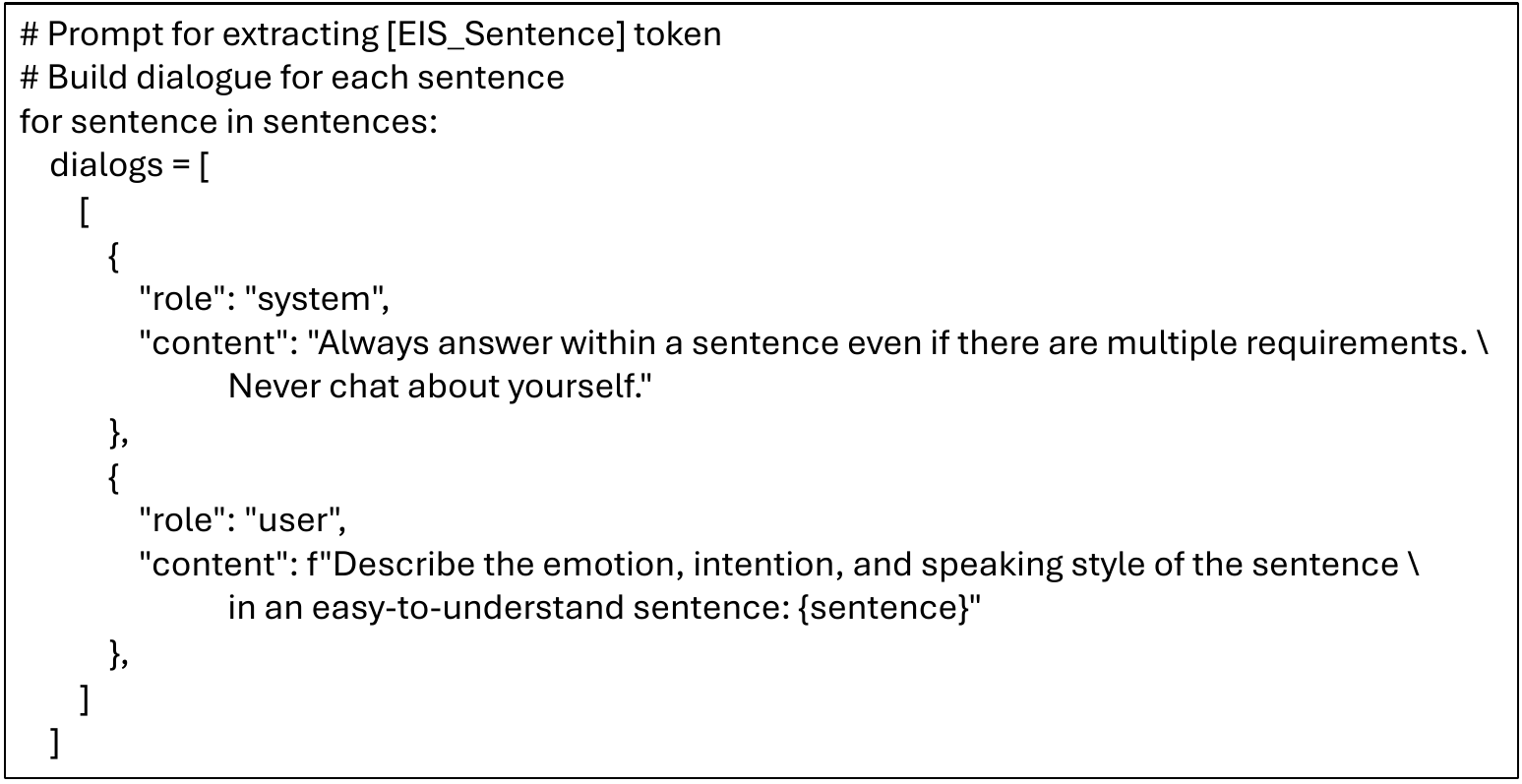}
        \caption{Prompts used for extracting [EIS\_Sentence] tokens.}
        \label{fig:prompt_sentence}
    \end{subfigure}
    \caption{Prompts used for extracting EIS tokens}
    \label{fig:prompts}
\end{figure*}

The main contributions encapsulate:
\begin{itemize}
\item We propose Llama-VITS model that utilizes the semantic understanding and expression capabilities of Llama2, offering equal or superior acoustic performance compared to baseline models, along with a significantly enhanced ability to understand and express semantics and emotions. 
\item Through empirical analysis, we demonstrate that global tokens in Llama-VITS provide more significant improvements than sequential tokens, contrasting with observations in BERT-based TTS models.
% \item We introduce a new evaluation framework that incorporates ESMOS, a new subjective metric to assess emotional expression, with traditional objective metrics for acoustic performance of synthesized speech. 
\item We quantitatively verified our findings using both subjective and objective metrics. 
\end{itemize}

Our code, models, audio demos, and the filtered single female speaker emotional dataset EmoV\_DB\_bea\_sem are available at \url{https://github.com/xincanfeng/vitsgpt.git}.

\section{Related Work}
% % TTS has problems like: acoustic information + semantic information can help.
% % llama has some application in speech. show understanding and generation sessions. 
% TTS技术在不断的结构进化和调整中，在模仿声学特征方面已经达到很高的水平，但在语义和感情的理解和表达方面一直是难点。与此同时，BERT及其类似的模型通过在大量文本上的预训练，获得了深层次的理解语义的能力，因此有许多研究都用BERT-like的语言模型结合在TTS技术中来增强合成的语音中对语义和情感的理解和表达。然而，在TTS技术中结合GPT及其类似的语言模型的研究却非常有限。
TTS technology has significantly advanced in learning acoustic features through structural evolution. However, comprehending and conveying semantics remain challenging. Since BERT-like LMs have demonstrated profound capabilities in understanding semantics through extensive pre-training on vast text corpora, some studies have integrated BERT-like LMs with TTS technology to enhance synthesized speech. Nonetheless, research on incorporating GPT-like LMs within TTS technology is notably scarce.

\subsection{Text-To-Speech Models}
\label{sec:tts models}

TTS task aims to generate natural, fluent, and easily comprehensible speech. Traditional TTS systems, e.g., a Statistical Parametric Speech Synthesis (SPSS) system~\citep{taylor2009text}, usually comprise multiple distinct components. These include a frontend module that converts text into linguistic features (such as duration and pitch), an acoustic model that maps these linguistic features to acoustic features, and a vocoder responsible for generating speech waveforms from the acoustic features. Over the past decades, the complexity of traditional models has been notable, attributed to their reliance on manually engineered features and the intricate communication between modules.

Transitioning from Hidden Markov Models (HMM) based models~\citep{Alan2007Statistical}, through Deep Neural Networks (DNN) models~\citep{zen2013Statistical}, to Generative Adversarial Networks (GAN) based models~\citep{saito2017statistical}, there has been a notable enhancement in voice quality, yet the architectural complexity remains significant.

The advent of end-to-end TTS models marks a significant milestone, increasingly reducing the distinction between synthesized speech and human voice. End-to-end models are capable of transforming raw text directly into final speech output, which not only streamlines the structural complexity of TTS systems and facilitates easier deployment but also significantly reduces the dependency on manual feature engineering, simplifying the training process. Moreover, they notably enhance the naturalness and intelligibility of the speech, thereby becoming the predominant architecture in TTS models. For instance, Char2Wav~\citep{sotelo2017Char2Wav} introduces an attentive encoder-decoder framework for direct speech synthesis from text input. Tacotron~\citep{wang2017tacotron} undertakes training from the ground up and directly predicts linear spectrograms. Furthermore, the speech produced by Tacotron2~\citep{shen2018natural} closely mirrors the natural human voice.

In the realm of end-to-end TTS models, many have adopted a non-autoregressive architecture. This architecture enables parallel data processing, where the model's output generation does not depend on the output of the previous time step, thereby enhancing processing speed. It also circumvents the error accumulation issue inherent in traditional autoregressive models, which significantly boosts TTS performance. FastSpeech~\citep{ren2019fastspeech} and its variants exemplify this trend. FastSpeech employs a transformer-based architecture to generate mel-spectrograms in parallel. Building on FastSpeech, FastPitch~\citep{Adrian2021fastpitch} predicts pitch contours during inference, enabling the production of more expressive and high-quality speech. FastSpeech2~\citep{ren2022fastspeech} further incorporates explicit duration prediction and introduces pitch and energy as conditional inputs.

Previous non-autoregressive approaches typically involve distinct training phases for acoustic models and vocoders. VITS~\citep{kim2021conditional} introduces a more natural-sounding output compared to these two-stage systems through its one-stage parallel end-to-end architecture. Innovatively, VITS incorporates variational inference combined with normalizing flows and employs an adversarial training methodology. Due to VITS's exemplary performance across multiple benchmarks, we select it as the foundational TTS model for our system.

\subsection{Fine-tuning BERT-like LMs for TTS}
\label{sec:former bert}

% 虽然TTS模型在模仿声学特征方面越来越进步，但如果训练数据不够多，模型可能无法学到同一个输入在不同上下文中的情感，从而表达能力受限。因此研究者们想到利用BERT及其类似模型强大的迁移学习能力。最终，结合了经过预训练和微调的BERT-like模型的TTS系统有了更好的理解语义，并提高了生成语音的表现力。这是TTS技术的重要进步。
While TTS models have increasingly advanced in replicating acoustic features, insufficient training data can hinder the model's ability to learn the semantic nuances of the same input across different contexts, thus limiting its expressiveness. Consequently, researchers have turned to leveraging the transfer learning capabilities of BERT-like LMs. Ultimately, TTS systems that incorporate pre-trained and fine-tuned BERT-like LMs have achieved better understandings of semantics and enhanced generated speech, marking a significant advancement.

\citet{shinji2019pretrained} utilized a pre-trained BERT model as an auxiliary input to enhance a Tacotron2-based TTS system, resulting in improved speech naturalness. Similarly, \citet{yang2019pretrained} applied a pre-trained BERT model to achieve enhanced front-end accuracy. \citet{kenter2020improving} demonstrated that integrating a BERT model, pre-trained on extensive unlabeled data and fine-tuned for speech, into an RNN-based TTS system enhances prosody. \citet{kenter2020improving} specifically suggest updating the BERT's parameters during the training of their RNN-based speech synthesis model, emphasizing the critical role of fine-tuning the BERT component for optimal outcomes. As prompt tuning draws wide attention in guiding text or image generation, PromptTTS~\citep{guo2022PromptTTS} takes a prompt representation with both style and content descriptions from a BERT model as input to generate speech with precise style control and high speech quality. 
% Specifically, PromptTTS consists of a style encoder and a content encoder to extract the corresponding representations from the prompt, and a speech decoder to synthesize speech according to the extracted style and content representations. 

In particular, \citet{mukherjee2022Text} utilized a pre-trained BERT model to develop a text emotion classification model, employing the final hidden states of the initial [CLS] token as a comprehensive representation of the text. Researchers such as \citet{kenter2020improving, li2021improving, abbas2022expressive} have applied word-level BERT to capture the semantic and syntactic structure of sentences, thereby aiding TTS synthesis. \citet{li2023phonemelevel} introduced a phoneme-level BERT, designed with a preliminary task of predicting corresponding graphemes in addition to regular masked phoneme predictions, to enhance the naturalness of speech synthesized from out-of-distribution (OOD) texts.

However, despite BERT's acknowledged capacity to provide detailed word importance, syntactic and semantic insights, and general knowledge~\citep{shinji2019pretrained, kenter2020improving}, its effectiveness is constrained by the particularities of fine-tuning approaches. Furthermore, BERT's inherent non-generative nature might limit its ability to account for information outside the immediate sentence context.

\subsection{Integrating GPT-like LMs for TTS}
\label{sec: Llama2}

% 考虑到在语义的理解和表达上，BERT模型主要用于理解任务，而GPT模型不仅能理解文本、还能生成自然连贯的文本，而且GPT模型拥有更大的模型参数，特别擅长零样本（zero-shot）或少样本（few-shot）学习，可以直接应用于多种任务而完全或基本不需要微调或修改模型结构，所以研究者们也想到使用GPT来辅助TTS。
Considering semantic understanding and expression capabilities, BERT is primarily utilized for comprehension tasks. In comparison, GPT excels not only in understanding text but also in generating natural and coherent text. Moreover, with the larger model parameters, GPT is particularly adept at zero-shot or few-shot learning, enabling its direct application to various tasks with little to no need for fine-tuning or structural modifications. 
% Consequently, researchers have also considered leveraging GPT to enhance TTS systems.

% 但是利用GPT-like的模型来辅助TTS系统的语义理解的研究非常有限。已知的比如，\citet{stephenson2021alternate} discusses enhancing the prosody of speech synthesis using the GPT model by adjusting text input.这样的方法会限制TTS的应用，因为更多时候我们并不希望更改输入。In terms of evaluation, it only assesses changes in internal model metrics like Duration and Energy, lacking a quantitative assessment of the final output. Its results were also not notably significant. 
However, research on leveraging GPT-like models to aid TTS systems is very limited. \citet{stephenson2021alternate} explores the potential of improving speech synthesis naturalness by text input lookahead with GPT prediction. Such an approach potentially restricts TTS applications, as altering the input is often undesirable. 
% Regarding evaluation, the study primarily focuses on internal model metrics such as Duration and Energy, without providing a quantitative analysis of the final output quality. 
Furthermore, the findings were not verified by human subjective evaluation. 
\citet{saito2023chatgptedss} suggest employing ChatGPT\footnote{\url{https://openai.com/blog/chatgpt}} to aid in empathetic dialogue speech synthesis by extracting the context of conversations. They particularly instruct ChatGPT to produce three keywords that encapsulate the intention, emotion, and speaking Style of speech observed in the dialogue history. These keywords are subsequently utilized to train a speech synthesis model. However, due to the inaccessibility of ChatGPT to the public, the researchers resort to processing ChatGPT's outputs with BERT to extract embeddings. This approach essentially positions ChatGPT as an alternative to manual annotation, yet it does not delve into investigating ChatGPT's internal representations and their potential impact on speech-related tasks.

% In this paper, we chose Llama2 as the GPT-like language model to be integrated into our TTS system, driven by its technological superiority and promising applications. Firstly, Llama2 is one of the largest publicly available language models, comparable to closed-source models like GPT3.5\cite{openai2024gpt4} and PaLM (540B)\cite{chowdhery2022palm}, and significantly outperforms other open-source models like MPT\footnote{\url{https://www.databricks.com/blog/mpt-30b}} and Falcon\cite{almazrouei2023falcon} in benchmark tests. Secondly, Llama2's innovative architecture offers high security and is conducive to expanding downstream tasks\cite{touvron2023llama}.

% Concurrent works\cite{radhakrishnan2023whispering, zhang2023videollama} that incorporate Llama2 in speech and other modal tasks, as well as the rapid development of computing costs reduction of Llama2\footnote{\url{https://huggingface.co/4bit/Llama-2-70b-chat-hf}}, reaffirm Llama2's research value and the anticipation it garners in multimodal tasks.

In our study, we selected Llama2, a GPT-like LM, for integration into our TTS system, motivated by its technological advancements and potential for diverse applications. Llama2 stands out as one of the largest publicly accessible LMs, rivaling proprietary models such as GPT3.5~\citep{openai2024gpt4} and PaLM (540B)~\citep{chowdhery2022palm}, and surpasses other open-source alternatives like MPT\footnote{\url{https://www.databricks.com/blog/mpt-30b}} and Falcon~\citep{almazrouei2023falcon} in benchmark evaluations. Additionally, the novel architecture of Llama2 not only ensures enhanced security but also facilitates the extension of various downstream tasks~\citep{touvron2023llama}.

Related research that employs Llama2 in speech and other multimodal tasks~\citep{radhakrishnan2023whispering, zhang2023videollama}, coupled with the ongoing efforts to reduce computing costs associated with Llama2\footnote{\url{https://huggingface.co/4bit/Llama-2-70b-chat-hf}}, underscores the model's significant research interest and its promising prospects in multimodal applications.

\section{Methodology}
% figure: token extraction

We propose leveraging semantic embeddings derived from a GPT-like LM to improve TTS synthesis. In our work, Llama2 is employed as the GPT-like model, as elaborated in Section~\S\ref{sec: Llama2}, and VITS is utilized as the TTS model for generating audio from phoneme embeddings, as detailed in Section~\S\ref{sec:tts models}. In essence, we extract semantic embeddings $E_s$ from the final hidden layer of Llama2 and integrate them with the original acoustic text embeddings $E_a$ of VITS, forming enhanced text embeddings $E_{as}$ for speech synthesis. Specifically, either a global token or a sequence of tokens is used to encapsulate the semantic attributes of an input sentence for varying objectives. The distinctions between these two token types are further explicated in Section~\S\ref{sec:sem tokens}.

\subsection{Semantic Embeddings Derived from Llama2}
\label{sec:sem tokens}

% 对每一个输入的句子s，我们抽取s在Llama2输出前的最后一个隐藏层的信息，并采取不同的策略来制作不同的tokens作为这个句子的semantic embedding。
For each input sentence \(s\), we extract information from the final hidden layer before the output of Llama2. Different strategies are employed to create various tokens that serve as the semantic embedding for the sentence.

% 设 \(E_s\) 表示句子 \(s\) 的语义embedding，\(H_{\text{Llama}}^F(s)\) 表示Llama2模型对于句子 \(s\) 在最后一个隐藏层 \(F\) 的输出。则 \(E_s\) 可以表示为：
% \[E_s = H_{\text{Llama}}^F(s)\]
% 这里，\(H_{\text{Llama}}^F(s)\) 是一个向量，它捕获了句子 \(s\) 经过Llama2模型所有层的处理后，在最后一层的表示。这个表示包含了句子的语义信息，并在我们的Llama-VITS系统中作为语音合成的输入。
Let \(E_s\) denote the semantic embedding of sentence \(s\), and \(H_{\text{Llama}}^F(s)\) represent the output of the Llama2 model for sentence \(s\) at the final hidden layer \(F\). Therefore, \(E_s\) can be expressed as:
\begin{align}
E_s = H_{\text{Llama}}^F(s)
\end{align}
Here, \(H_{\text{Llama}}^F(s)\) is a vector that encapsulates the semantic representation of sentence \(s\) after processing through all layers of the Llama2, culminating in the final layer. 

\paragraph{Formulation for Global Tokens}

% 我们一共尝试了五种global tokens，分别为[AVE], [PCA], [LAST], [EIS\_Word]和[EIS\_Sentence]，每种策略都使用一个token来表示输入句子的全局语义特征。

% 在[AVE]策略中，通过取句子 \(s\) 的输出向量中所有tokens的平均值来作为semantic token，用以下公式表示：
% \[E_s^{\text{AVE}} = \frac{1}{n} \sum_{i=1}^{n} H_{\text{Llama}}^F(s, i)\]
% 其中，\(E_s^{\text{AVE}}\) 表示用[AVE]策略得到的句子\(s\)的semantic token，\(H_{\text{Llama}}^F(s, i)\) 代表Llama2模型对于句子 \(s\) 在最后一个隐藏层 \(F\) 的输出中第 \(i\) 个token的表示。句子 \(s\) 包含 \(n\) 个tokens。

% 在[PCA]策略中，我们对句子 \(s\) 的输出向量应用PCA抽取主成分，并根据原始数据的数值范围对PCA结果的均值进行缩放来得到semantic token。其中，缩放是为了使PCA处理后的数据在数值上与原始数据保持一致的规模，从而保留语义信息在数值表示上的相对重要性。这个策略可以用以下公式表示：
% \[E_s^{\text{PCA}} = \text{PCA\_rescale}\left( H_{\text{Llama}}^F(s) \right)\]

% 在[LAST]策略中，我们直接取句子 \(s\) 的输出向量中的最后一个token来作为semantic token，用以下公式表示：
% \[E_s^{\text{LAST}} = H_{\text{Llama}}^F(s, n)\]
% 其中，\(H_{\text{Llama}}^F(s, n)\) 是指句子 \(s\) 经过Llama2模型所有层的处理后，在最后一层对应句子最后一个token的表示。

We explored five types of global tokens to represent the overarching semantic features of an input sentence, namely [AVE], [PCA], [LAST], [EIS\_Word], and [EIS\_Sentence], with each strategy employing a single token.

In the [AVE] strategy, the semantic token is derived by calculating the average of all tokens' output vectors for sentence \(s\), formulated as:
\begin{align}
E_s^{\text{AVE}} = \frac{1}{n} \sum_{i=1}^{n} H_{\text{Llama}}^F(s, i)
\end{align}
Here, \(E_s^{\text{AVE}}\) denotes the semantic token obtained using the [AVE] strategy, and \(H_{\text{Llama}}^F(s, i)\) represents the output of the \(i\)th token of sentence \(s\) at the final hidden layer \(F\) of Llama2, with \(s\) comprising \(n\) tokens.

For the [PCA] strategy, we apply Principal Component Analysis to the output vectors of sentence \(s\) to extract principal components and rescale the mean of the PCA results according to the original data's value range. This rescaling ensures that the PCA-processed data maintains a scale consistent with the original data, preserving the relative importance of semantic information numerically. Formulated as:
\begin{align}
E_s^{\text{PCA}} = \text{PCA\_rescale}\left( H_{\text{Llama}}^F(s) \right)
\end{align}

In the [LAST] strategy, the semantic token is obtained by selecting the last token from the output vector of sentence \(s\), as shown in the formula:
\begin{align}
E_s^{\text{LAST}} = H_{\text{Llama}}^F(s, n)
\end{align}
where \(H_{\text{Llama}}^F(s, n)\) refers to the representation of the last token of sentence \(s\) after processing through all layers of Llama2 at the final layer.

% 在[EIS\_Word]和[EIS\_Sentence]策略中，不同于以上使用句子本身表示的策略，我们直接获取Llama2对句子s的理解来作为这个句子的语义表示。具体来说，我们分别用\ref{fig:prompt_word}和\ref{fig:prompt_sentence}所示的prompt来得到Llama2模型对句子s在Emotion,Intention和speaking Style三个方面的理解u，并计算这个输出的理解的表示的均值来作为semantic token。在[EIS\_Word]策略中，我们引导Llama2分别三个词语来形容Emotion,Intention和speaking Style，最终的semantic token公式如下：
% \begin{align}
% E_s^{\text{EIS\_Word}} &= \frac{1}{m} \Bigl[ \sum_{i} H_{\text{Llama}}^{F}(u_{\text{E}},i) \nonumber\\
% &+ \sum_{j} H_{\text{Llama}}^{F}(u_{\text{I}},j) + \sum_{k} H_{\text{Llama}}^{F}(u_{\text{S}},k) \Bigr]
% \end{align}
% 其中$u_{\text{E}}, u_{\text{I}}, u_{\text{S}}$分别是Llama2输出的用三个词语来表达的对原句子的理解在最后一个隐藏层中的表示，$i,j,k$分别表示每个输出的词语的token，$m$是这些token的总数。

% 在[EIS\_Sentence]中，我们引导Llama2用一个简单且易于理解的句子u来形容它对输入句子s的Emotion,Intention和speaking Style方面的理解，最终的semantic token为：
% \begin{align}
% E_s^{\text{EIS\_Sentence}} = \frac{1}{m} \sum_{i=1}^{n} H_{\text{Llama}}^F(u_{\text{EIS}}, i)
% \end{align}
% 其中$u_{\text{EIS}}$是Llama2输出的用一个句子来表达的对原句子的理解在最后一个隐藏层中的表示，$m$是这个句子表示的所有token的总数。

In the [EIS\_Word] and [EIS\_Sentence] strategies, unlike the above approaches that utilize the sentence itself for representation, we derive the semantic representation of sentence \(s\) based on Llama2's comprehension $u$. Adapted from~\citet{saito2023chatgptedss}'s practice, we employ prompts as illustrated in \ref{fig:prompt_word} and \ref{fig:prompt_sentence}, respectively, to obtain Llama2's understanding of sentence \(s\) in terms of Emotion, Intention, and speaking Style, denoted as \(u\), and calculate the average of this understanding's representation to serve as the semantic embedding. 

In the [EIS\_Word] strategy, Llama2 is prompted to describe Emotion, Intention, and speaking Style with three separate words, resulting in the following formula for the final semantic token:
\begin{align}
E_s^{\text{EIS\_Word}} =& \frac{1}{m} \Bigl[ \sum_{i} H_{\text{Llama}}^{F}(u_{\text{E}},i) \nonumber\\
&+ \sum_{j} H_{\text{Llama}}^{F}(u_{\text{I}},j) + \sum_{k} H_{\text{Llama}}^{F}(u_{\text{S}},k) \Bigr]
\end{align}
where \(u_{\text{E}}, u_{\text{I}}, u_{\text{S}}\) are the representations of Llama2's output expressing the sentence's Emotion, Intention, and speaking Style at the final hidden layer, respectively, with \(i, j, k\) indicating the tokens of each output word, and \(m\) being the total number of these tokens.

In the [EIS\_Sentence] strategy, Llama2 is guided to describe its understanding of the input sentence's Emotion, Intention, and speaking Style with an easy-to-understand sentence, leading to the following formula for the final semantic token:
\begin{align}
E_s^{\text{EIS\_Sentence}} = \frac{1}{m} \sum_{i=1}^{m} H_{\text{Llama}}^F(u_{\text{EIS}}, i)
\end{align}
where \(u_{\text{EIS}}\) is the representation of Llama2's output expressing the understanding of the original sentence at the final hidden layer, and \(m\) is the total number of tokens in this sentence representation.

\paragraph{Formulation for Sequential Tokens}
% 在采用sequential tokens策略时，我们关注于使用句子的全部tokens来表示其语义信息。与global tokens策略不同，sequential tokens策略不仅包括了基于文本的表示，还考虑了基于音素的表示，以更好地适应TTS模型可能对声学特征的关注。以下是这两种策略下semantic token的数学表示：

% ### [TEX] 策略
% 在[TEX]策略下，我们直接使用句子 \(s\) 的文本形式的全部tokens来表示其语义信息。如果句子 \(s\) 在Llama2模型的最后一个隐藏层 \(F\) 的输出由 \(n\) 个tokens组成，则semantic token \(T_s^{\text{TEX}}\) 可以表示为一个序列：
% \[T_s^{\text{TEX}} = \{H_{\text{Llama}}^F(s, 1), H_{\text{Llama}}^F(s, 2), \ldots, H_{\text{Llama}}^F(s, n)\}\]

% ### [PHO] 策略
% 在[PHO]策略下，我们考虑句子 \(s\) 的音素形式的全部tokens来表示其语义信息。这里，\(s_{\text{pho}}\) 表示句子 \(s\) 转换为音素表示的形式，如果 \(s_{\text{pho}}\) 在Llama2模型的最后一个隐藏层 \(F\) 的输出由 \(m\) 个tokens组成，则semantic token \(T_s^{\text{PHO}}\) 可以表示为一个序列：
% \[T_s^{\text{PHO}} = \{H_{\text{Llama}}^F(s_{\text{pho}}, 1), H_{\text{Llama}}^F(s_{\text{pho}}, 2), \ldots, H_{\text{Llama}}^F(s_{\text{pho}}, m)\}\]

% 在这两种策略中，\(H_{\text{Llama}}^F(s, i)\) 和 \(H_{\text{Llama}}^F(s_{\text{pho}}, i)\) 分别代表Llama2模型对于句子 \(s\) 的文本形式和音素形式在最后一个隐藏层 \(F\) 的第 \(i\) 个token的输出。这样的表示允许TTS模型利用句子的完整语义信息，无论是基于文本还是基于音素的输入。

In the implementation of sequential tokens strategies, we focus on utilizing the complete set of tokens from the input sentence to represent its semantic information. Unlike the global token approaches, sequential tokens strategies encompass representations based on either text or phonemes, aiming to better align with the TTS model's potential emphasis on acoustic features. The mathematical representations for these two strategies are as follows:

% ### [TEX] Strategy
Under the [TEX] strategy, we directly employ all tokens from the textual form of sentence \(s\) to represent its semantic information. If the output of sentence \(s\) at the final hidden layer \(F\) of Llama2 consists of \(n\) tokens, then the semantic token \(T_s^{\text{TEX}}\) is represented as a sequence:
\begin{align}
E_s^{\text{TEX}} = \{H_{\text{Llama}}^F(s, 1), H_{\text{Llama}}^F(s, 2), \ldots, H_{\text{Llama}}^F(s, n)\}
\end{align}

% ### [PHO] Strategy
In the [PHO] strategy, we consider the complete set of tokens from the phonemic form. Here, \(s_{\text{pho}}\) denotes the phonemic representation of sentence \(s\). If the output of \(s_{\text{pho}}\) at the final hidden layer \(F\) of Llama2 comprises \(m\) tokens, then the semantic token \(T_s^{\text{PHO}}\) is represented as a sequence:
\begin{align}
E_s^{\text{PHO}} =& \{H_{\text{Llama}}^F(s_{\text{pho}}, 1), \nonumber\\
&H_{\text{Llama}}^F(s_{\text{pho}}, 2), \ldots, H_{\text{Llama}}^F(s_{\text{pho}}, m)\}
\end{align}

In both strategies, \(H_{\text{Llama}}^F(s, i)\) and \(H_{\text{Llama}}^F(s_{\text{pho}}, i)\) respectively represent the outputs of the \(i\)th token of sentence \(s\) in its textual and phonemic forms at the final hidden layer \(F\) of Llama2. This representation allows the TTS model to leverage the complete semantic information of a sentence, whether based on text or phonemes.

\subsection{Fusing Semantic Embedding with Acoustic Embedding}

% 为了将从Llama2提取的语义嵌入\(E_s\)的维度与VITS原本的声学嵌入\(E_a\)的维度统一，我们可以使用线性投影。设 \(E_s\) 的原始维度为 \(d_{\text{Llama}}\)，我们希望将其投影到与VITS文本嵌入相同的维度 \(d_{\text{VITS}}\)。这个线性投影可以通过一个线性变换矩阵 \(W\) 实现，其中 \(W\) 的维度为 \(d_{\text{VITS}} \times d_{\text{Llama}}\)。因此，投影后的语义嵌入 \(E_s'\) 可以通过以下公式计算：
% \begin{align}
% E_s' = W \cdot E_s
% \end{align}
To align the dimensions of semantic embedding extracted from Llama2, denoted as \(E_s\), with the acoustic embeddings\ from VITS, denoted as \(E_a\), we employ a linear projection. The original dimension of \(E_s\), \(d_{\text{Llama}}\), is projected to match the dimension of VITS acoustic embedding, \(d_{\text{VITS}}\), using a linear transformation matrix \(W\) of dimensions \(d_{\text{VITS}} \times d_{\text{Llama}}\). The projected semantic embedding, \(E_s'\), is calculated as follows:
\begin{align}
E_s' = W \cdot E_s
\end{align}

\paragraph{Fusing Global Embedding with Acoustic Embedding}
% 为了得到融合了语义与声学信息的嵌入\(E_{as}\)，对于global tokens，我们用简单相加的方式，把上述维度统一后的global embeddings与VITS中的text embedding融合，公式如下：
% \begin{align}
% E_{as} = E_a + E_s'
% \end{align}
To obtain an embedding \(E_{as}\) that integrates both semantic and acoustic information, for global tokens, we simply add the dimensionally unified global embedding to VITS's acoustic embedding, as shown in the equation:
\begin{align}
E_{as} = E_a + E_s'
\end{align}

\paragraph{Fusing Sequential Embeddings to Enhance Text Embeddings}
% 我们使用Scaled Dot Product Attention将sequential embedding与VITS原始的acoustic embedding进行融合来得到\(E_{as}\)，以下数学公式可以表达这个计算过程：

% 给定：\(q\) 为acoustic embedding in VITS，维度为 \([b, t, d]\); \(k\) 和 \(v\) 为semantic embedding from Llama2，维度也为 \([b, t, d]\); \(b\) 为batch size，\(t\) 为序列长度，\(d\) 为embedding维度; \(\text{temperature}\) 为缩放因子。

% 首先，计算attention scores \(A\)：
% \[A = \frac{q \cdot k^T}{\text{temperature}}\]

% 其中，\(k^T\) 表示 \(k\) 的转置，使得 \(k\) 从 \([b, t, d]\) 转置为 \([b, d, t]\)，以便进行矩阵乘法。结果 \(A\) 的维度为 \([b, t, t]\)。

% 如果存在source mask或target mask，将会应用mask操作，将mask位置的attention scores设置为一个非常小的值（例如，\(-6e4\)），以便在后续的softmax步骤中这些位置几乎不贡献权重。

% 接着，对attention scores应用softmax函数并进行dropout处理，得到最终的attention weights \(W_{\text{attn}}\)：
% \[W_{\text{attn}} = \text{Dropout}(\text{Softmax}(A, \text{dim}=-1))\]

% 最后，使用attention weights对 \(v\) 进行加权求和，得到输出 \(O\)：
% \[O = W_{\text{attn}} \cdot v\]

% 输出 \(O\) 的维度为 \([b, t, d]\)，与 \(q\) 的维度相同。这个输出可以被视为融合了semantic information的text embedding，可以直接用于后续的TTS模型处理。

% 总结，这个融合过程通过计算text embedding \(q\) 和semantic embedding \(k, v\) 之间的scaled dot-product attention，然后使用这个attention来加权求和 \(v\)，从而获得融合了语义信息的新的embedding表示。

We utilize the Scaled Dot-Product Attention mechanism to merge sequential embeddings with VITS's original acoustic embedding to gain enhanced embedding \(E_{as}\), which can be described by the following mathematical formulas:

First, calculate the attention scores \(A\):
\begin{align}
A = \frac{q \cdot k^T}{\gamma}
\end{align}
where \(q\) is the acoustic embedding \(E_a\) in VITS with dimensions \([b, t, d]\); \(k\) and \(v\) denotes the semantic embedding \(E_s'\) from Llama2, also with dimensions \([b, t, d]\); \(b\) is the batch size, \(t\) is the sequence length, and \(d\) is the embedding dimension; \(\gamma\) is temperature for scaling. \(k^T\) denotes the transpose of \(k\), transforming \(k\) from \([b, t, d]\) to \([b, d, t]\) for matrix multiplication. The resulting \(A\) has dimensions \([b, t, t]\).

If a source mask or target mask is present, a masking operation is applied, setting the attention scores at masked positions to a very low value (e.g., \(-6e4\)) to nearly eliminate their weight contribution in the subsequent softmax step.

Next, apply the softmax function and dropout to the attention scores, obtaining the final attention weights \(W_{\text{attn}}\):
\begin{align}
W_{\text{attn}} = \text{Dropout}(\text{Softmax}(A))
\end{align}

Finally, the output \(E_{as}\) is calculated by weighting \(v\) with the attention weights:
\[E_{as} = W_{\text{attn}} \cdot v\]

The output \(E_{as}\), viewed as text embedding fused with semantic information, has dimensions \([b, t, d]\) that match those of \(q\). 

\begin{table*}[htbp]
    \centering
  % \caption{Results on full LJSpeech, bold text show the best result in each model.}
    \begin{subtable}{.86\linewidth}
    \centering
    \resizebox{\linewidth}{!}{
    \small
    {
    \renewcommand{\arraystretch}{0.9}
    \begin{tabular}{llllcccc}
    \toprule
    \multicolumn{8}{c}{\textbf{full LJSpeech}} \\
    \midrule
    \multirow{3}[6]{*}{\textbf{Model}} & \multicolumn{3}{l}{\textbf{Semantic Token}} & \multicolumn{4}{c}{\textbf{Evaluation 100k-step (batchsize=64)}} \\
\cmidrule{2-8}          & \multirow{2}[4]{*}{\textbf{Type}} & \multirow{2}[4]{*}{\textbf{Fuse}} & \multicolumn{1}{l}{\multirow{2}[4]{*}{\textbf{Token}}} & \multirow{2}[4]{*}{\textbf{UTMOS}} & \multirow{2}[4]{*}{\textbf{MCD}} & \multicolumn{2}{c}{\textbf{ASR}} \\
\cmidrule{7-8}          &       &       &       &       &       & \textbf{CER} & \textbf{WER} \\
    \midrule
    ORI-VITS & -     & -     & -     & 4.19 ± 0.05 & 7.32 ± 0.61 & 6.2   & 16.5 \\
    \midrule
    \multirow{3}[6]{*}{BERT-VITS} & \multicolumn{1}{p{2.085em}}{gl} & add   & CLS   & 4.10 ± 0.06 & 7.38 ± 0.60 & 6.0     & 16.5 \\
\cmidrule{2-8}          & \multicolumn{1}{l}{\multirow{2}[4]{*}{seq}} & \multirow{2}[4]{*}{att} & BERT\_TEX & \textbf{4.22 ± 0.05} & \textbf{7.27 ± 0.61} & \textbf{5.9} & 15.9 \\
\cmidrule{4-8}          &       &       & BERT\_PHO & 4.08 ± 0.08 & 7.28 ± 0.62 & \textbf{5.9} & \textbf{15.7} \\
    \midrule
    \multirow{7}[14]{*}{\textbf{Llama-VITS}} & \multicolumn{1}{l}{\multirow{5}[10]{*}{gl}} & \multirow{5}[10]{*}{add} & AVE   & \textbf{4.21 ± 0.05} & 7.30 ± 0.66 & 5.9   & 16 \\
\cmidrule{4-8}          &       &       & PCA   & 4.19 ± 0.05 & \textbf{7.23 ± 0.61} & \textbf{5.8} & \textbf{15.8} \\
\cmidrule{4-8}          &       &       & LAST  & \textbf{4.21 ± 0.05} & 7.39 ± 0.63 & \textbf{5.8} & 16.2 \\
\cmidrule{4-8}          &       &       & EIS\_Word & 4.16 ± 0.06 & 7.32 ± 0.62 & \textbf{5.8} & 16.1 \\
\cmidrule{4-8}          &       &       & EIS\_Sentence & \textbf{4.21 ± 0.04} & 7.26 ± 0.64 & 5.9   & 16.2 \\
\cmidrule{2-8}          & \multirow{2}[4]{*}{seq} & \multirow{2}[4]{*}{att} & TEX   & 4.13 ± 0.06 & 7.48 ± 0.68 & 6.0     & 16 \\
\cmidrule{4-8}          &       &       & PHO   & 4.19 ± 0.05 & 7.33 ± 0.65 & 6.0     & 16.2 \\
    \bottomrule
    \end{tabular}}
    }
    \label{tab:ljs24}%
\end{subtable}%

\begin{subtable}{.86\linewidth}
    \centering
    % \caption{Results on 1-hour LJSpeech. Notation is the same as that in Table \ref{tab:ljs24}.}
    \resizebox{\linewidth}{!}{
    \small
    {
    \renewcommand{\arraystretch}{0.9}
    \begin{tabular}{llllcccc}
    \toprule
    \multicolumn{8}{c}{\textbf{1-hour LJSpeech}} \\
    \midrule
    \multirow{3}[6]{*}{\textbf{Model}} & \multicolumn{3}{l}{\textbf{Semantic Token}} & \multicolumn{4}{c}{\textbf{Evaluation 100k-step (batchsize=64)}} \\
\cmidrule{2-8}          & \multirow{2}[4]{*}{\textbf{Type}} & \multirow{2}[4]{*}{\textbf{Fuse}} & \multicolumn{1}{l}{\multirow{2}[4]{*}{\textbf{Token}}} & \multirow{2}[4]{*}{\textbf{UTMOS}} & \multirow{2}[4]{*}{\textbf{MCD}} & \multicolumn{2}{c}{\textbf{ASR}} \\
\cmidrule{7-8}          &       &       &       &       &       & \textbf{CER} & \textbf{WER} \\
    \midrule
    ORI-VITS & -     & -     & -     & 4.02 ± 0.08 & 7.47 ± 0.63 & 6.0     & 16.1 \\
    \midrule
    \multirow{3}[6]{*}{BERT-VITS} & \multicolumn{1}{p{2.165em}}{gl} & add   & CLS   & 4.02 ± 0.07 & \textbf{7.39 ± 0.62} & \textbf{6.0} & \textbf{16.3} \\
\cmidrule{2-8}          & \multicolumn{1}{l}{\multirow{2}[4]{*}{seq}} & \multirow{2}[4]{*}{att} & BERT\_TEX & 3.91 ± 0.08 & 7.54 ± 0.60 & 6.2   & 16.6 \\
\cmidrule{4-8}          &       &       & BERT\_PHO & \textbf{4.05 ± 0.07} & 7.40 ± 0.65 & 6.2   & \textbf{16.3} \\
    \midrule
    \multirow{7}[14]{*}{\textbf{Llama-VITS}} & \multicolumn{1}{l}{\multirow{5}[10]{*}{gl}} & \multirow{5}[10]{*}{add} & AVE   & \textbf{4.10 ± 0.07} & 7.37 ± 0.63 & \textbf{6.0} & \textbf{16.4} \\
\cmidrule{4-8}          &       &       & PCA   & 4.04 ± 0.06 & 7.39 ± 0.60 & 6.4   & 16.7 \\
\cmidrule{4-8}          &       &       & LAST  & 3.98 ± 0.08 & 7.38 ± 0.62 & 6.1   & 16.6 \\
\cmidrule{4-8}          &       &       & EIS\_Word & 3.99 ± 0.08 & 7.37 ± 0.60 & 6.2   & 16.6 \\
\cmidrule{4-8}          &       &       & EIS\_Sentence & 4.01 ± 0.08 & \textbf{7.36 ± 0.59} & \textbf{6.0} & \textbf{16.4} \\
\cmidrule{2-8}          & \multirow{2}[4]{*}{seq} & \multirow{2}[4]{*}{att} & TEX   & 4.02 ± 0.08 & 7.55 ± 0.62 & 6.2   & 16.7 \\
\cmidrule{4-8}          &       &       & PHO   & 3.95 ± 0.08 & 7.42 ± 0.57 & 6.7   & 17.3 \\
    \bottomrule
    \end{tabular}}
    }
  \label{tab:ljs1}%
\end{subtable}%

\begin{subtable}{.86\linewidth}
  \centering
  \resizebox{\linewidth}{!}{
    \small
    {
    \renewcommand{\arraystretch}{0.9}
    \begin{tabular}{llllccccc}
    \toprule
    \multicolumn{9}{c}{\textbf{EmoV\_DB\_bea\_sem}} \\
    \midrule
    \multirow{3}[6]{*}{\textbf{Model}} & \multicolumn{3}{l}{\textbf{Semantic Token}} & \multicolumn{5}{c}{\textbf{Evaluation 150k-step (batchsize=16)}} \\
\cmidrule{2-9}          & \multirow{2}[4]{*}{\textbf{Type}} & \multirow{2}[4]{*}{\textbf{Fuse}} & \multicolumn{1}{l}{\multirow{2}[4]{*}{\textbf{Token}}} & \multirow{2}[4]{*}{\textbf{ESMOS}} & \multirow{2}[4]{*}{\textbf{UTMOS}} & \multirow{2}[4]{*}{\textbf{MCD}} & \multicolumn{2}{c}{\textbf{ASR}} \\
\cmidrule{8-9}          &       &       &       &       &       &       & \textbf{CER} & \textbf{WER} \\
    \midrule
    ORI-VITS & -     & -     & -     & 3.06 ± 0.08 & 3.61 ± 0.08 & 7.06 ± 1.19 & 4.5   & 18.5 \\
    \midrule
    \multirow{3}[6]{*}{BERT-VITS} & \multicolumn{1}{p{2.915em}}{gl} & add   & CLS   & \textbf{3.02 ± 0.07} & 3.50 ± 0.11 & \textbf{7.11 ± 1.13} & 5.4   & 20.5 \\
\cmidrule{2-9}          & \multicolumn{1}{l}{\multirow{2}[4]{*}{seq}} & \multirow{2}[4]{*}{att} & BERT\_TEX & 2.92 ± 0.08 & \textbf{3.61 ± 0.18} & 7.21 ± 1.15 & \textbf{4.4} & \textbf{18.7} \\
\cmidrule{4-9}          &       &       & BERT\_PHO & 2.96 ± 0.08 & 3.50 ± 0.15 & 7.21 ± 1.26 & 5.0     & 19.6 \\
    \midrule
    \multirow{7}[14]{*}{\textbf{Llama-VITS}} & \multicolumn{1}{l}{\multirow{5}[10]{*}{gl}} & \multirow{5}[10]{*}{add} & AVE   & 3.01 ± 0.08 & 3.60 ± 0.13 & 7.15 ± 1.18 & 4.6   & 18.3 \\
\cmidrule{4-9}          &       &       & PCA   & -     & 3.57 ± 0.13 & \textbf{7.13 ± 1.24} & 4.8   & 19.2 \\
\cmidrule{4-9}          &       &       & LAST  & -     & 3.55 ± 0.11 & \textbf{7.13 ± 1.17} & \textbf{4.3} & \textbf{17.4} \\
\cmidrule{4-9}          &       &       & EIS\_Word & -     & \textbf{3.61 ± 0.12} & 7.20 ± 1.23 & 4.7   & 19.6 \\
\cmidrule{4-9}          &       &       & EIS\_Sentence & -     & 3.55 ± 0.12 & \textbf{7.13 ± 1.23} & 5.0     & 18.5 \\
\cmidrule{2-9}          & \multirow{2}[4]{*}{seq} & \multirow{2}[4]{*}{att} & TEX   & \textbf{3.22 ± 0.07} & 3.40 ± 0.21 & 7.21 ± 1.35 & 4.5   & 18.3 \\
\cmidrule{4-9}          &       &       & PHO   & 2.98 ± 0.07 & 3.52 ± 0.15 & 7.27 ± 1.30 & 4.7   & 18.5 \\
    \bottomrule
    \end{tabular}}
    }
  \label{tab:emovdb}%
\end{subtable}%
\caption{Results on full LJSpeech, 1-hour LJSpeech, and EmoV\_DB\_bea\_sem dataset, respectively. Bold text show the best result in each model. Note that, for global tokens extracted from Llama2, ESMOS metric is only evaluated on [AVE] token to save human cost.}
\label{tab:all results}
\end{table*}

\section{Experiments}
\subsection{Experimental Settings}
We propose Llama-VITS which uses semantic tokens derived from Llama2 to enhance acoustic embedding in VITS for better TTS performance. To show the effectiveness of our method, we experimented with two baseline models. 
In the ORI-VITS baseline, we use the original VITS without external semantic information. 
In the BERT-VITS baseline, we extract various semantic tokens according to former research introduced in Section \S~\ref{sec:former bert}. Specifically, we use the [CLS] token of BERT as the global token. To form the baseline of the sequential token in BERT, we use all the tokens in the sentence trained by text or phoneme, named [BERT\_TEX] and [BERT\_PHO], respectively. 
In our proposed Llama-VITS, we derive global token [AVE], [LAST], [PCA], [EIS\_Word], and [EIS\_Sentence], and sequential tokens [TEX] and [PHO] from Llama2, corresponding to those in BERT-VITS.  

We use Llama2 (13b) to generate semantic embeddings of dimension 5120. [CLS] and [BERT\_TEX] tokens are extracted from BERT-base-uncased model which has a parameter size of 110M that generates token embedding of 768 dimensions.  [BERT\_PHO] token is extracted from BERT-x-phone-base model whose parameter size is 88M to generate token embedding of 768 dimensions.

\paragraph{Global Token Extraction}
In our proposed Llama-VITS, global strategy [LAST] only uses the last token in the final hidden layer of Llama2 for each sentence.  [AVE] uses the average of all tokens for each sentence.  [PCA] uses the concatenation of all tokens whose dimension was reduced by Principal Component Analysis (PCA). [EIS\_Word] and [EIS\_Sentence] use the average of tokens for an answer, which is formed in three words or a sentence by prompts shown in Figure~\ref{fig:prompts}, to describe the Emotion, Intention, and speaking Style of the transcript. 

In BERT-VITS baseline, global strategy [CLS] only uses the first token from the BERT-base-uncased model for each input sentence. 

\paragraph{Sequential Token Extraction}
In our proposed Llama-VITS, sequential strategy [TEX] concatenates the sequence of tokens in a sentence generated by Llama2 using text input. [PHO] concatenates the sequence of tokens of a sentence generated by Llama2 using phonemic input. 

In the baseline BERT-VITS, sequential strategy [BERT\_TEX] concatenates all the tokens in a sentence extracted from BERT-base-uncased model. [BERT\_PHO] concatenates all the tokens in a sentence extracted from BERT-x-phone-base model. 

\paragraph{Datasets}
We utilized full LJSpeech, 1-hour LJSpeech, and EmoV\_DB\_bea\_sem dataset for experimental verification. LJSpeech\footnote{\url{https://keithito.com/LJ-Speech-Dataset/}} comprises 24 hours recorded of English speech by single female speaker, where we evaluate how the embeddings extracted from Llama2 can help improve the speech naturalness. Besides full LJSpeech dataset, we also randomly filtered 1-hour LJSpeech which contains only 1-hour records as an ablation study to show how dataset size influences. EmoV\_DB\footnote{\url{https://www.openslr.org/115/}}~\citeplanguageresource{adigwe2018emotional} is a database of emotional speech that contains data for male and female actors in English and French. EmoV\_DB covers 5 emotion classes, amused, angry, disgusted, neutral, and sleepy. To factor out the effect of different speakers, we filtered the original EmoV\_DB dataset into the speech of a specific female English speaker, bea. Then we use Llama2 to predict the emotion label of the transcript chosen from the above 5 emotion classes, and select the audio samples which has the same predicted emotion. The filtered dataset contains 22.8-min records for training. We named the filtered dataset EmoV\_DB\_bea\_sem and investigated how the semantic embeddings from Llama2 behave in naturalness and expressiveness on it. Please refer to Appendix A~\ref{app:dataset} for more dataset statistics. 

\paragraph{Implementation, Hyper-parameters, Training}
Our Llama-VITS system was built on the VITS~\citep{kim2021conditional} framework using its original implementation\footnote{\url{https://github.com/jaywalnut310/vits}}, augmented with semantic embeddings derived from Llama2~\citep{touvron2023llama} using its original implementation\footnote{\url{https://github.com/facebookresearch/llama}}. For training LJSpeech, we use the public configs in the original implementation of VITS. For EmoV\_DB\_bea\_sem, we use the same config as LJSpeech but changed batch size from 64 to 16 since this dataset is much smaller. Besides implementing our proposed Llama-VITS, we extracted corresponding semantic tokens [CLS], [BERT\_TEX] from BERT uncased base model\footnote{\url{https://huggingface.co/bert-base-uncased}} and [BERT\_PHO] from BERT pre-trained on phoneme\footnote{\url{https://huggingface.co/vinai/xphonebert-base}} for comparison. 

In comparing the experimental results, we choose 100k-step results on both full LJSpeech and 1-hour LJSpeech datasets since they are rather large. On EmoV\_DB\_bea\_sem, we used the pre-trained checkpoint of LJSpeech on 100k-step and compare the fine-tuning results on EmoV\_DB\_bea\_sem at 150k-step since it is rather small.  

\paragraph{Evaluation Metrics}
Both subjective and objective metrics are implemented for a comprehensive evaluation.
% \begin{itemize}
%     \item Mean Opinion Score (MOS), higher is better.
%     \item Emotion Similarity Mean Opinion Score (ESMOS), higher is better.
%     \item UTokyo-SaruLab Mean Opinion Score (UTMOS), higher is better.
%     \item Mel-Cepstral Distortion (MCD), lower is better.
%     \item Automatic Speech Recognition (ASR), lower is better. 
% \end{itemize}
In subjective evaluation, we conduct Emotion Similarity Mean Opinion Score (ESMOS)~\citep{zhu2023multispeaker} experiments to evaluate emotion similarity for EmoV\_DB\_bea\_sem. In the subjective evaluation, we compared [AVE], [TEX] and [PHO] strategies in our Llama-VITS with the corresponding token [CLS], [BERT\_TEX] and [BERT\_PHO] extracted from different BERT models and the baseline ORI-VITS who does not contain semantic tokens, with the ground truth samples GT. 

% For LJSpeech, we randomly choose 150 samples from the total 500 test samples of the original implementation in \citep{kim2021conditional} and invite 3 native English speakers to participate in the listening tests. Each participant is asked to choose the naturalness in a 5 scale: Excellent 5, Good 4, Fair 3, Poor 2, Bad 1 for the MOS score. 
In evaluating ESMOS, we randomly chose 5 samples from the total 51 test samples proportionally divided by us and received 100 test results from different speakers on Amazon Mechanical Turk. The result significance level is thus 500. Each participant is asked to give a score on emotion similarity compared with ground truth in a 5-scale: Excellent Match 5, Good Match 4, Fair Match 3, Poor Match 2, Bad Match 1\footnote{Note that in the ESMOS experiments, participants are asked to ignore the speakers' voice, style, and audio quality and only consider the emotiveness of the speech.}. 

In objective evaluation, we utilize UTokyo-SaruLab Mean Opinion Score (UTMOS)~\citeplanguageresource{saeki2022utmos}, Mel-Cepstral Distortion (MCD), and speech recognition performance measured by Character Error Rate (CER) and Word Error Rate (WER). UTMOS is a MOS prediction network using speech samples from previous Blizzard Challenges and Voice Conversion Challenges, which has reached the best performance in VoiceMOS Challenge 2022. We evaluate objective intelligibility by using Whisper-large~\citep{radford2022robust}. 
For calculating UTMOS, we use the implementation in SpeechMOS\footnote{\url{https://github.com/tarepan/SpeechMOS}}.
For calculating MCD and ASR, we use the evaluation implementation\footnote{\url{https://github.com/espnet/espnet}} of ESPnet \citeplanguageresource{hayashi2020espnettts, hayashi2021espnet2tts}.

\section{Experimental Results}
We evaluated our proposed Llama-VITS along with baselines ORI-VITS and BERT-VITS models on three distinct datasets: the full LJSpeech, the 1-hour LJSpeech, and EmoV\_DB\_bea\_sem. The experimental outcomes provide a comprehensive understanding of the model performance and the impact of semantic tokens selection. A summary of these results is articulated below and can be referenced in Table \ref{tab:all results}. 

\subsection{Results on full LJSpeech}
The ORI-VITS baseline, achieving a UTMOS of \(4.19 \pm 0.05\), an MCD of \(7.32 \pm 0.61\), a CER of \(6.2\), and a WER of \(16.5\).

Enhancements were observed with the BERT-VITS baseline. Specifically, BERT-VITS with [BERT\_TEX] semantic tokens demonstrated superior performance in UTMOS (\(4.22 \pm 0.05\)) and MCD (\(7.27 \pm 0.61\)), indicating improved speech quality and reduced mel-cepstral distortion. Additionally, a reduced CER of \(5.9\) and WER of \(15.9\) were noted, highlighting enhanced automatic speech recognition accuracy.

Our proposed Llama-VITS, integrating various global and sequential semantic tokens, displayed competitive performance. The [PCA] strategy stood out, achieving an MCD of \(7.23 \pm 0.61\), indicating optimal mel-cepstral distortion. The [EIS\_Sentence], [AVE], and [LAST] tokens yielded a top-tier UTMOS of \(4.21 \pm 0.04/0.05\), underscoring their effectiveness in enhancing perceived speech quality.

\subsection{Results on 1-hour LJSpeech}
In the more challenging 1-hour LJSpeech dataset, all models experienced a slight performance decrease, an expected outcome given the reduced training data size.

BERT-VITS baseline with [CLS] tokens exhibited notable MCD performance (\(7.39 \pm 0.62\)), while the [BERT\_PHO] excelled in UTMOS (\(4.05 \pm 0.07\)), reflecting enhanced speech naturalness and reduced mel-cepstral distortion.

Llama-VITS with [AVE] tokens achieved the highest UTMOS (\(4.10 \pm 0.07\)), while [EIS\_Sentence] tokens resulted in the most favorable MCD (\(7.36 \pm 0.59\)), illustrating the model's versatility and efficacy in different token configurations.

\subsection{Results on EmoV\_DB\_bea\_sem}
On this even more challenging dataset, 
% ORI-VITS baseline achieves a ESMOS of \(3.06 \pm 0.08\), a UTMOS of \(3.61 \pm 0.08\), an MCD of \(7.06 \pm 1.19\), a CER of \(4.5\), and a WER of \(18.5\).
a small improvement observed in BERT-VITS only exists in the [BERT\_TEX] with a CER of \(4.4\). 

While our proposed Llama-VITS displayed notable enhancements. The [TEX] strategy achieves an ESMOS of \(3.22 \pm 0.07\), indicating much more emotiveness. The [LAST] yielded the best performance on CER of \(4.3\) and WER of \(17.4\), other strategies also perform better than or comparable to BERT-VITS, underscoring its effectiveness in enhancing perceived speech expressiveness.

\subsection{Analysis}
Speaking of the strengths of different tokens, BERT-based tokens generally contribute to improving MCD and ASR scores, indicating the enriched semantic understanding translated to speech quality.
Tokens of Llama-VITS exhibited a balanced performance across all metrics, with specific token configurations excelling in particular aspects. For instance, [PCA] token emerged as a strong contender in reducing MCD, [AVE] enhanced the UTMOS scores, [TEX] had superior performance to improve ESMOS score. 

In individual comparisons, Llama-VITS's five global tokens generally outperformed BERT-VITS on the UTMOS metric for naturalness. In the ESMOS metric for emotional expression, Llama-VITS's two sequential tokens also generally surpassed BERT-VITS, particularly the [TEX] token. Therefore, we can infer that GPT-like LMs may have greater potential for TTS tasks than BERT-like models. 

Further, our results reflect different patterns of gains from GPT-like and BERT-like models in TTS tasks. For instance, in the UTMOS naturalness metric, Llama-VITS's global tokens often outperformed sequential tokens, which is the opposite for BERT-VITS; in the ESMOS emotion metric, Llama-VITS's sequential token [TEX] significantly outperformed other tokens, while for BERT-VITS, global tokens performed better. 

Overall, Llama-VITS showed a different pattern in UTMOS compared to BERT-VITS, and superior performance in ESMOS. These results highlight the potential for further exploration of semantic token types and fusion methods to achieve more significant enhancements in speech synthesis, particularly in scenarios constrained by limited and complex training data.

\section{Discussions}
% 在这一章节中，我们讨论现有结果的可能因素、以及未来的工作方向。
In this section, we discuss factors influencing current outcomes. Based on this discussion, we also point out the directions for future work in Appendix~\ref{app:future}. 

\subsection{GPT-like vs BERT-like}
% The main reason that we consider is that LJSpeech is a neutral dataset lacking expressiveness, significantly limiting the performance of semantic understanding and consequently the effectiveness of semantic tokens provided by Llama2. 

% To validate how different datasets influence, we specifically filtered one hour of the LJSpeech dataset, i.e., 1-hour LJSpeech, and filtered the original EmoV\_DB dataset to match LJSpeech in terms of being a single female speaker in English, i.e., EmoV\_DB\_bea\_sem. This made the two filtered datasets similar in data size and speaker characteristics, except for the emotional expression, enabling us to compare that the semantic understanding in LJSpeech is constrained by its mode of neutral emotional expression. 

% 首先，从我们现有的实验可以看出，即便是在对Llama没有任何微调的情况下，Llama-VITS在情感表达上的性能远远超出BERT-VITS和ORI-VITS，这为之后探索不同的TTS中的情感表达任务提供了可能。
Initial observations from our experiments indicate that, even without any fine-tuning of Llama2, Llama-VITS significantly outperforms both BERT-VITS and ORI-VITS in terms of emotional expressiveness. This finding opens up avenues for future research into emotive TTS tasks.

Furthermore, a comparison between BERT-VITS and Llama-VITS highlights their distinct performance traits. BERT-VITS, leveraging deep contextual embeddings, provides profound semantic insights yet encounters challenges in customization and adaptability across a range of TTS tasks. Conversely, Llama-VITS can provide a more versatile and adaptable approach, with its array of token types demonstrating particular advantages across various evaluation metrics.

\subsection{Semantic Token Strategy}
The varying effectiveness of distinct semantic tokens underscores the importance of careful selection and integration tailored to the particular goals of TTS systems. Optimizing the type of token and method of fusion can be instrumental in enhancing aspects such as speech naturalness, emotional expressiveness, Mel Cepstral Distortion (MCD), or Automatic Speech Recognition (ASR) performance.

\section{Conclusion}
In summary, this study exemplifies a significant stride towards optimized TTS synthesis by integrating semantic tokens, leveraging the strengths of Llama-VITS. Our findings, validated by comprehensive experiments on the LJSpeech and EmoV\_DB\_bea\_sem datasets, underscore the pivotal role of semantic embeddings in enhancing speech quality, naturalness, and emotiveness. The adaptability and efficacy of Llama-VITS, especially, open new vistas for customized and context-sensitive TTS applications. 

\section{Limitations}

Compared with our baseline which uses different BERT models, we only tested our method using Llama2. As \citet{kenter2020improving} indicate for their BERT-based TTS model, small BERT models work better than big ones, but the parameter size of our proposed GPT-based TTS influence is yet studied by our research. Although BERT-based TTS models are normally finetuned on speech tasks to provide more explicit acoustic information for TTS, we didn't try designing prompts to generate acoustic features and only studied how general semantic information can help. Our experiments were conducted only on clean datasets with limited size, and the effect on more complex datasets is to be further explored. The integration of Llama2's embeddings introduces additional computational costs, potentially limiting real-time applications. 

\section{Acknowledgements}
This research was conducted with the support of team members who contributed to varying extents. Particular gratitude is extended to Koichi Miyazaki for his sharing regarding foundational knowledge, his assistance in implementing the subjective evaluation and precious advice. We are also deeply appreciative of Masato Murata, Katsuhiko Yamamoto, and Li Li for their insightful suggestions to enrich the presentation of our paper and code.

% \section{Optional Supplementary Materials}
% Appendices or supplementary material (software and data) will be allowed ONLY in the final, camera-ready version, but not during submission, as papers should be reviewed without the need to refer to any supplementary materials.
% Each \textbf{camera ready} submission can be accompanied by an appendix usually being included in a main PDF paper file, one \texttt{.tgz} or \texttt{.zip} archive containing software, and one \texttt{.tgz} or \texttt{.zip} archive containing data.
% TODO: privide the filtered dataset made by us. 

\nocite{*}
\section{Bibliographical References}\label{sec:reference}

\bibliographystyle{lrec-coling2024-natbib}
\bibliography{lrec-coling2024-example}

\section{Language Resource References}
\label{lr:ref}
\bibliographystylelanguageresource{lrec-coling2024-natbib}
\bibliographylanguageresource{languageresource}

\section{Appendix A. Dataset Statistics}
\label{app:dataset}

We have summarized the dataset statistics in the table below.

\begin{table*}[htbp]
  \centering
  \begin{tabularx}{\textwidth}{>{\hsize=9.5\hsize}X>{\hsize=4\hsize}X>{\hsize=4\hsize}X>{\hsize=4\hsize}X>{\hsize=4\hsize}X>{\hsize=10\hsize}X} 
    \toprule
    \textbf{Dataset} & \textbf{Train} & \textbf{Dev} & \textbf{Test} & \textbf{Total} & \textbf{Emotions} \\
    \midrule
    full LJSpeech & 22.8hour & 10.6min & 54.5min & 24hour & neutral \\
    1-hour LJSpeech & 1hour & 10.6min & 54.5min & 2hour & neutral \\
    EmoV\_DB\_bea\_sem & 15min & 4min & 3.8min & 22.8min & neutral, amused, angry, disgusted, sleepy \\
    \bottomrule
  \end{tabularx}
  \caption{Dataset statistics. Note that, the LJSpeech dataset, a public-domain speech dataset recorded by a single female speaker, necessitated the filtration of the original EmoV\_DB dataset to isolate recordings by a single female speaker, Bea, resulting in our creation of the EmoV\_DB\_bea dataset. This process was undertaken to facilitate a more accurate comparison. Additionally, the EmoV\_DB\_bea dataset underwent further refinement by selecting audio recordings whose expressed emotions aligned with those identified in the transcripts by Llama2, culminating in the EmoV\_DB\_bea\_sem dataset. This dataset enables the training of models to recognize and learn emotional expressions.}
  \label{tab:datasets}%
\end{table*}

\section{Appendix B. Future Work}
\label{app:future}

Given the promising potential of incorporating advanced semantic tokens to transform the TTS synthesis landscape, enhancing speech naturalness, emotiveness, and overall quality, future research could focus on investigating various token types, refining fusion strategies, and assessing performance across an expanded range of datasets and contexts.

Furthermore, with the rapid evolution of LM architectures and advancements in quantization techniques, a wider array of LM options has become feasible. For instance, Mixtral~\cite{jiang2024mixtral}, a Sparse Mixture of Experts (SMoE) LM, is considered comparable in capability to Llama2.

Revisiting the initial rationale for selecting GPT-like models, given that the effectiveness of Llama-VITS in improving speech synthesis performance has been established, it is conceivable to leverage the more adaptable fine-tuning or prompting features of GPT-like models over BERT-like models. This approach could explore utilizing GPT-like models for enhanced control over TTS synthesis tasks and refining fine-tuning or prompting methodologies.

Looking ahead, examining the dynamics of various TTS models when combined with semantic tokens could also be pivotal in unlocking the full potential of semantic-enhanced TTS systems.

\end{document}